\documentclass[sn-basic]{sn-jnl}% Basic Springer Nature Reference Style/Chemistry Reference Style
\usepackage{natbib}
\usepackage{algorithm}
\usepackage{algpseudocode}
\usepackage{caption}
\usepackage{subcaption}

\jyear{2023}%

\theoremstyle{thmstyleone}%
\theoremstyle{thmstyletwo}%

\theoremstyle{thmstylethree}%

\usepackage{booktabs}

\begin{document}

\title[adaptNMT: an open-source, language-agnostic NMT development environment]{adaptNMT: an open-source, language-agnostic development environment for Neural Machine Translation}


\author*[1,2]{\fnm{Séamus} \sur{Lankford}}\email{seamus.lankford@mtu.ie}

\author[2]{\fnm{Haithem} \sur{Afli}}\email{haithem.afli@mtu.ie}

\author[1]{\fnm{Andy} \sur{Way}}\email{andy.way@adaptcentre.ie}

\affil*[1]{\orgdiv{ADAPT Centre}, \orgname{Dublin City University}, \orgaddress{\city{Dublin}, \country{Ireland}}}

\affil[2]{\orgdiv{ADAPT Centre}, \orgname{Munster Technological University}, \orgaddress{\city{Cork}, \country{Ireland}}}


\abstract{adaptNMT streamlines all processes involved in the development and deployment of RNN and Transformer neural translation models. As an open-source application, it is designed for both technical and non-technical users who work in the field of machine translation. Built upon the widely-adopted OpenNMT ecosystem, the application is particularly useful for new entrants to the field since the setup of the development environment and creation of train, validation  and test splits is greatly simplified. Graphing, embedded within the application, illustrates the progress of model training, and SentencePiece is used for creating subword segmentation models. Hyperparameter customization is facilitated through an intuitive user interface, and a single-click model development approach has been implemented. Models developed by adaptNMT can be evaluated using a range of metrics, and deployed as a translation service within the application. To support eco-friendly research in the NLP space, a green report also flags the power consumption and kgCO\textsubscript2 emissions generated during model development. The application is freely available.\footnote{\url{http://github.com/adaptNMT}}}




\maketitle

\section{Introduction}\label{sec1}

Explainable Artificial Intelligence (XAI) \citep{gunning2019xai, arrieta2020explainable} seeks to ensure that the results of AI solutions are easily understood by humans. It is against this backdrop that adaptNMT has been developed to afford users a form of {\em Explainable Neural Machine Translation (XNMT)}. The stages involved in a typical NMT process are broken down into a series of independent steps including environment setup, dataset preparation, training of subword models, parameterizing and training of main models, evaluation and deployment. This modular approach has created an effective NMT model development process for both technical and less technical practitioners in the field. Given the  environmental impact of building and running of large AI models \citep{strubell-etal-2019-energy,henderson2020towards,jooste-etal-2022-knowledge}, we also compute carbon emissions in a `green report', primarily as an information aid, but hopefully as  a way to encourage reusable and sustainable model development.

An important part of this research involves developing applications and models to address the challenges of language technology. It is hoped that such work will be of particular benefit to newcomers to the field of Machine Translation (MT) and in particular to those who wish to learn more about NMT.

In order to have a thorough understanding of how NMT models are trained, the individual components and the mathematical concepts underpinning both RNN- and Transformer-based models are explained and illustrated in this paper. The  application is built upon OpenNMT \citep{klein2017opennmt} and subsequently inherits all of its features. Unlike many NMT toolkits, a CLI (command line interface) approach is not used. The interface is designed and fully implemented in Google Colab.\footnote{\url{colab.research.google.com}} For an educational setting, and indeed for research practitioners, a Colab cloud-hosted\footnote{\url{cloud.google.com}} solution is often more intuitive to use. Furthermore, the training of models can be viewed and controlled using the Google Colab mobile app which is ideal for  builds with long run times. GUI controls, also implemented within adaptNMT, enable the customization of all key parameters required when training NMT models.

The application can be run in local mode enabling existing infrastructure to be utilised, or in hosted mode which allows for rapid scaling of the infrastructure. A deploy function allows for the immediate deployment of trained models. 

This paper is organized by initially presenting background information on NMT and related work on system-building environments in Section \ref{related}. This is followed by a detailed description of the adaptNMT architecture and its key features in Section \ref{arch}. An empirical evaluation of models is carried out in Section \ref{sec:exp}. The system is discussed in Section \ref{disc} before drawing conclusions and describing future work in Section \ref{concl}. For newcomers to the field, we suggest going straight to Section \ref{arch} to examine the platform's capabilities, and then discovering more about the various components and their statistical underpinning in Section \ref{related}. This can be followed by the remaining sections in their logical sequence. 

\section{Neural Networks for MT}\label{related}

\subsection{Recurrent Neural Network Architectures}
 
 Recurrent Neural Networks (RNNs)~\citep{sennrich2016edinburgh, sennrich2019revisiting, araabi2020optimizing} are often used for the tasks of Natural Language Processing (NLP), speech recognition and MT. RNNs, such as Long Short-Term Memory (LSTM)~\citep{hochreiter1997long}, were designed to support sequences of input data. LSTM models use an encoder-decoder architecture which enables variable length input sequences to predict variable length output sequences. This architecture is the cornerstone of many complex sequence prediction problems such as speech recognition and MT.

RNN models enable previous outputs to be used as inputs through the use of hidden states. In the context of MT, such neural networks were ideal due to their ability to process inputs of any length. In the initial stages of NMT, the RNN encoder-decoder framework was adopted and variable-length source sentences were encoded as fixed-length vectors~\citep{cho2014properties, sutskever2014sequence}. 
An improvement upon the basic RNN approach was proposed in \citet{bahdanau2014neural} which enhanced translation performance of the basic encoder-decoder architecture by replacing fixed-length vectors with variable-length vectors. A bidirectional RNN was now employed to read input sentences in the forward direction to produce forward hidden states while also producing backward hidden states by reading input sentences in the reverse direction. This development enabled neural networks to more accurately process long sentences, which previously had served as bottlenecks to performance, given their tendency to `forget' words in long input sequences which are `too far away' from the current word being processed. 

More importantly, \citet{bahdanau2014neural} introduced the concept of `attention' to the basic RNN architecture, similar in spirit and intention to `alignments' in the forerunner to NMT, Statistical MT \citep{och-ney-2003-systematic}. In attention-augmented NMT, the system could now pay special heed to the most relevant other source-sentence words and use them as contextual clues when considering how best to select the most appropriate target words(s) for translationally ambiguous words in the same string.

\subsection{Transformer Architecture}

Following the introduction of the attention mechanism, a natural line of investigation was to see whether attention could do most of the heavy lifting of translation by itself. Accordingly, \citet{vaswani2017attention} proposed that  “attention is all you need” in their `Transformer architecture', which has achieved state-of-the-art (SOTA) performance on many NLP benchmarks by relying solely on an attention mechanism,  removing recurrence and convolution, while allowing the use of much simpler feed-forward neural networks.

\begin{figure}[htb]
    \centering
    \includegraphics[width=7cm]{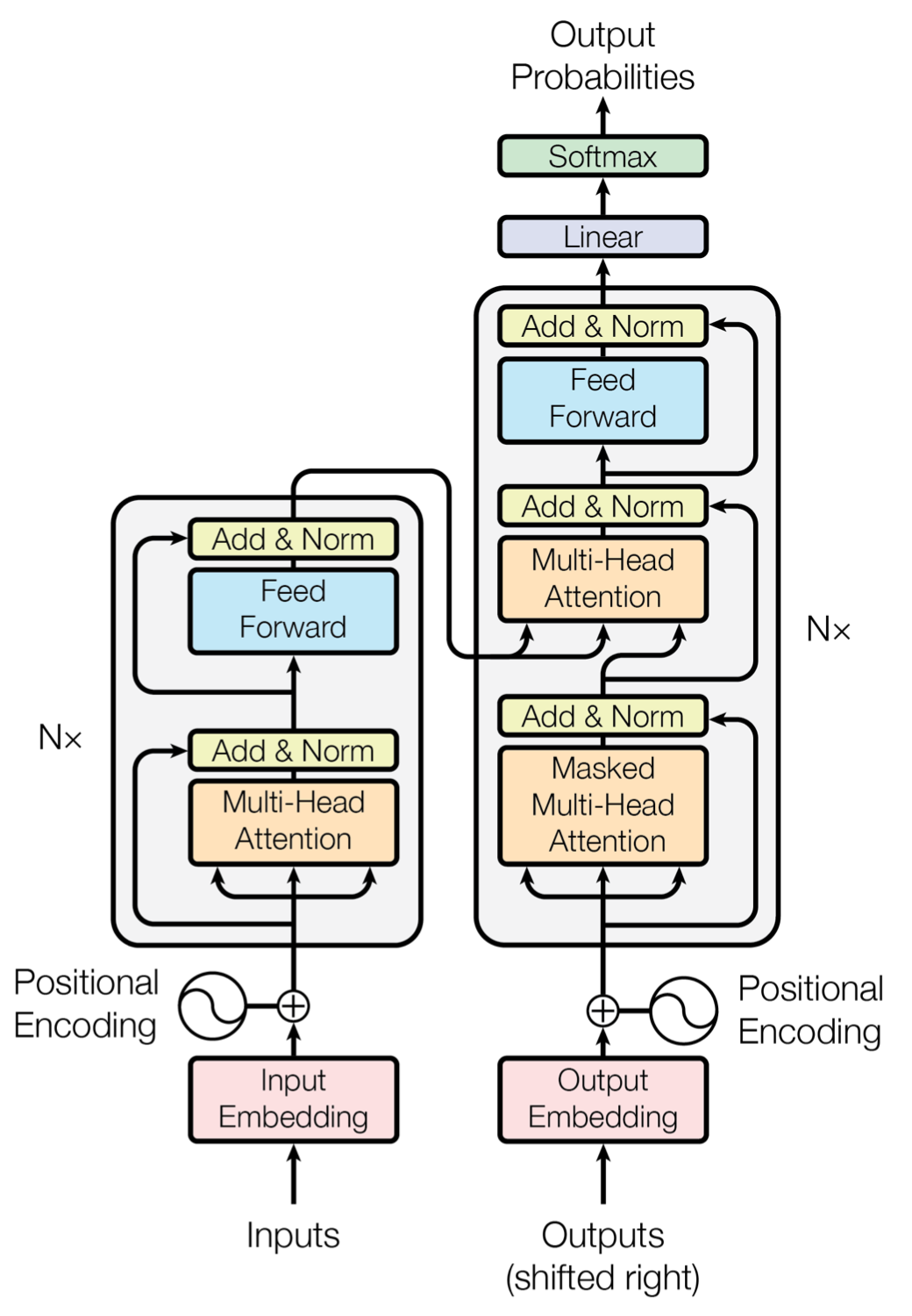}
    \caption{ \footnotesize {The Transformer architecture using an encoder-decoder~\citep{vaswani2017attention}. The encoder maps an input sequence to the decoder. The decoder generates a new output by combining the encoder output with the decoder output from the previous step.} }
    \label{fig:translayers}
\end{figure}

This approach follows an encoder-decoder structure, and allows models to develop a long memory which is particularly useful in the  area of language translation. The task of the encoder is to map an input sequence to a sequence of continuous representations, which is then passed to a decoder to generate an output sequence by using the output of the encoder together with the decoder output from the previous time step. Both the encoder and decoder each consist of a stack of 6 identical layers, whose structure  is illustrated in Figure \ref{fig:translayers}. In  the encoder, each layer is composed of two sub-layers: a multi-head self-attention mechanism and a fully connected feed-forward network. In the case of the decoder, there are three sub-layers: one which takes the previous output of the decoder stack, another which implements a multi-head self-attention mechanism, and the final layer which implements a fully connected feed-forward network.

\subsection{Attention}

\begin{figure}[h!]
    \centering
%    \captionsetup{justification=centering}
    \includegraphics[width=2.5cm]{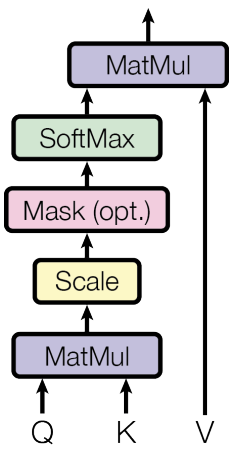}
    \caption{ \footnotesize { Multi-Head Attention in the Decoder~\citep{vaswani2017attention}. In the decoder, a multi-head layer receives queries from the previous decoder sublayer, and the keys and values from the encoder output. The decoder can now attend to all words in the input sequence.} }
    \label{fig:attention}
\end{figure}

As illustrated in Figure \ref{fig:attention}, the attention function can be described as mapping a query and a set of key-value pairs to an output, where the query, keys, values, and output are all vectors. The output is computed as a weighted sum of the values, where the weight assigned to each value is computed by a compatibility function of the query with the corresponding key,  as shown in Equation \eqref{eqn8}.

The query, keys and values used as inputs to the the attention mechanism are different projections of the same input sentence (`self-attention') and capture the relationships between the different words of the same sentence.

Both a scaled dot-product attention and a multi-head attention are used in the Transformer architecture. With scaled dot-product attention, a dot product is initially computed for each query $q$ with all of the keys $k$. Subsequently, each result is divided by $\sqrt{d_k}$ and a Softmax function is applied. The process leads to the weights which are used to scale the values, $v$.

The Softmax function allows us to perform multiclass classification which makes it a good choice in the final layer of neural network-based classifiers. The function forces the outputs of the neural network to a total sum to 1, which can be viewed as a probability distribution across multiple classes. Therefore, Softmax is the ideal choice as the output activation function, given that NMT is essentially a multiclass classification problem where the output classes represent the words within the vocabulary.

Computations performed by scaled dot-product attention can be efficiently applied on the entire set of queries simultaneously. To achieve this, the matrices, $Q$, $K$ and $V$, are supplied as inputs to the attention function:

\begin{equation}\label{eqn8}
attention(Q,K,V)=softmax(QK^T/\sqrt{d_k})V
\end{equation}

\subsection{NMT}\label{subsec2}
While much research effort concentrates on creating new SOTA NMT models, excellent descriptions of the technology are also available within the literature for those starting out in the field, or for those with a less technical background \citep{forcada2017making,wayBlooms}.

The availability of large parallel corpora has enabled NMT to develop high-performing MT models. Breakthrough performance improvements in the area of MT have been achieved through research efforts focusing on NMT~\citep{bahdanau2014neural} but the advent of the Transformer architecture has greatly improved MT performance. Consequently, SOTA performance has been attained on multiple language pairs~\citep{bojar-etal-2017-findings, bojar-etal-2018-findings, lankford2021transformer, lankford2022human, lankford2022lrec}. 

Similar to many deep-learning approaches, NMT development is underpinned by the mathematics of probability. At a fundamental level, the goal is to predict the probabilistic distribution $P (y\vert x)$ given a dataset $D$, where $x$ represents the source input sentence and $y$ represents the target output sentence. 

Supervised training of an NMT model develops the model weights by comparing the predicted $P (y\vert x)$ with the correct $y$ sentences of the training dataset, $D\textsubscript{Train}$. In evaluating the performance of an NMT model, automatic evaluation results are determined when the predicted $P (y\vert x)$ sentences are compared with the correct $y$ sentences of the test dataset, $D\textsubscript{Test}$. 

In adopting a deep learning paradigm, MT inherits the mathematical first principles which are inherent to this approach. To understand these principles, the manner in which neural networks model a conditional distribution is outlined. Furthermore, the encoder-decoder mechanism used for training NMT models is presented in the modelling subsection, and model optimization using training objectives is outlined in the learning subsection. Finally, the mathematics of how translated sentences are generated is explored in the inference subsection.

\subsubsection{Modelling}
In NMT, sentence-level translation is modelled using input and output sentences as sequences. Using this approach, an NMT model implements a sequence-to-sequence model with a given source sentence,  $x = (x_{1},...,x_{s})$ generating a target sentence $y = (y_{1},...,y_{t})$.

In effect, such a sequence-to-sequence NMT model acts as a conditional language model. The decoder within the model predicts the next word of the target sentence $y$, while such predictions are conditioned on the source sentence $x$.

By applying the chain rule, a model’s prediction (i.e. translation $y$ of length $T$) maximizes the probability $P (y\vert x)$ identified in Equations \eqref{eqn1} and \eqref{eqn2}:

\begin{equation} \label{eqn1}
P(y\vert x)=P(y_1\vert x)P(y_2\vert y_1,x)P(y_3\vert y_1, y_2,x)P(y_T\vert y_1,...,y_{T-1},x)
\end{equation}

\begin{equation} \label{eqn2}
P(y\vert x)=\prod_{t=1}^{T}P(y_t\vert y_1,...,y_{T-1},x)  
\end{equation}

\begin{figure}[htb]
    \centering
    \includegraphics[width=8cm]{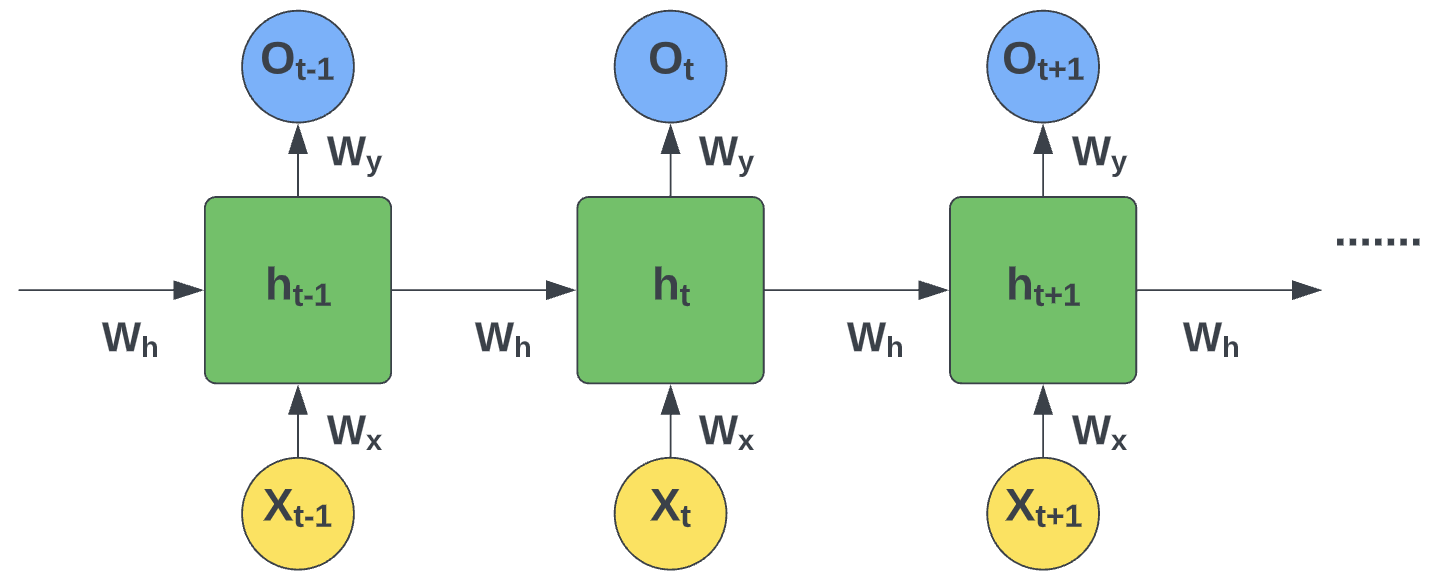}
    \caption{\footnotesize {Neurons within an RNN. At the input side, the neuron's input at time $t$ is a function of the encoded word (i.e. input vector $x_t$) and a hidden state vector $h_{t-1}$ which contains the previous sequence. The output generated by the neuron is represented by the vector $O_t$.}}
    \label{fig:neurons}
\end{figure}

Prior to Transformer, encoder-decoder models that incorporate RNNs were the most common method of representing text sequences in NMT.  RNNs are networks which accumulate information composed of similar units repeated over time. In  NMT, a primary function of the RNN encoder is that it encodes text, i.e. it turns text into a numeric representation. Neurons within an RNN are illustrated in Figure \ref{fig:neurons}.

\begin{figure}[htb]
    \centering
    \includegraphics[width=12cm]{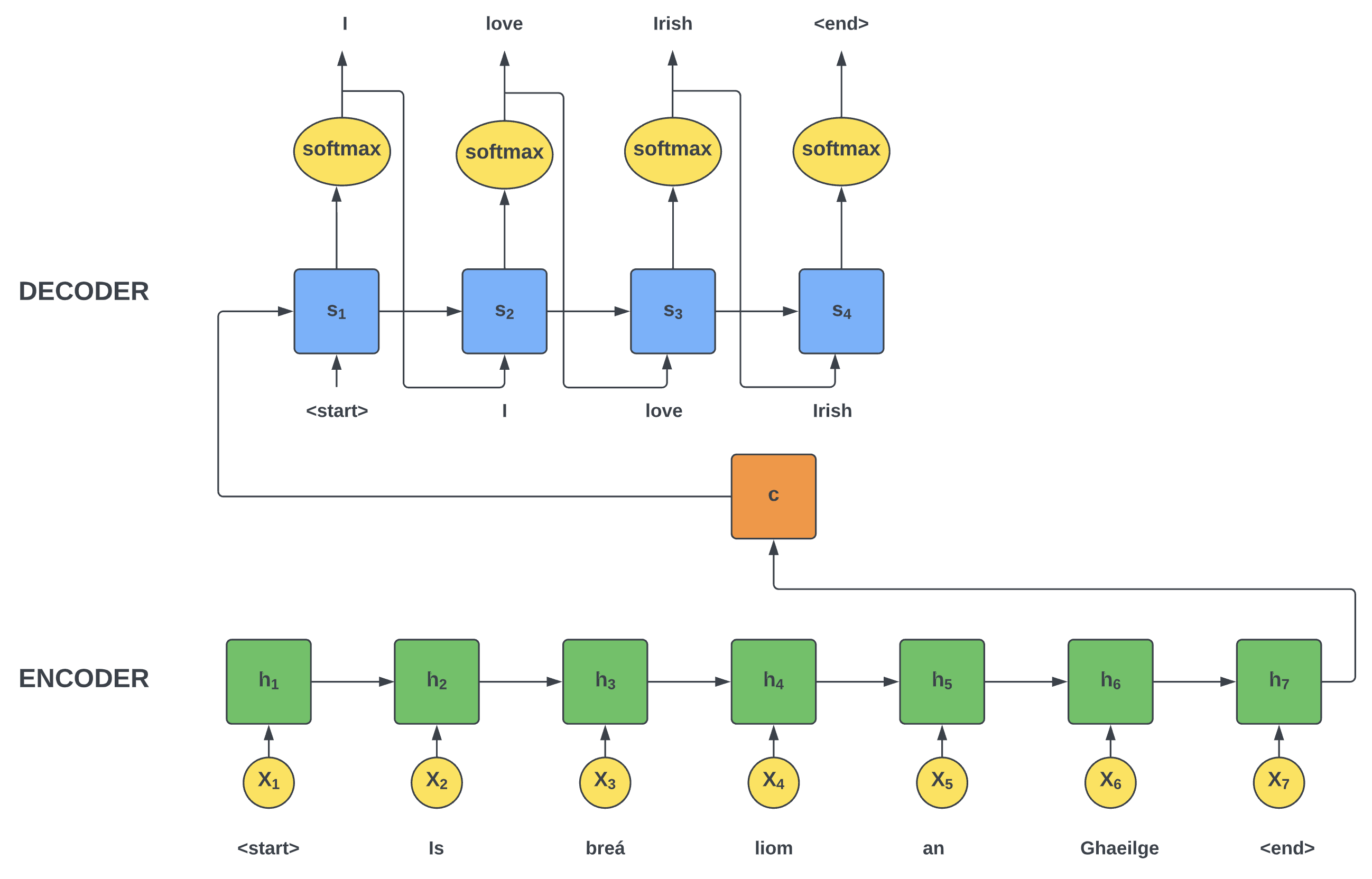}
    \caption{ \footnotesize { Encoder-decoder architecture. The encoder encodes the entire input sequence into a fixed-length \textit{context vector}, $c$, by processing input time steps. The function of the decoder is to read this \textit{context vector} while stepping through output time steps.} }
    \label{fig:encdec}
\end{figure}

Decoders unfold the vector representing the sequence state and return text. An important distinction between an encoder and a decoder is illustrated in Figure \ref{fig:encdec}, where it can be seen that both the encoder hidden state and the output from the previous decoding state are required by the decoder.

To kick-start processing of the decoder, a special token \textless start\textgreater ~is used since there is no previous output. The calculations carried out by the encoder are summarized in Equation \eqref{eqn3}:

\begin{equation} \label{eqn3}
h_t=RNN_{ENC}(x_t,h_{t-1})  
\end{equation}
\\
The $RNN_{ENC}$ function is iteratively applied over the input sequence to generate the final encoder state, $h_s$ which is fed to the decoder. The complete source sentence is effectively represented by $h_s$. The decoder within the model predicts the next word of the target sentence y, while such predictions are conditional on the source sentence $x$. 

The RNN decoder, $RNN_{DEC}$, creates a state vector $s_t$ by compressing the decoding history ${y_0,…,y_{t-1}}$ which is described in Equation \eqref{eqn4}. The distribution of target tokens is predicted by a classification layer which typically uses the Softmax activation function. 

\begin{equation} \label{eqn4}
s_t=RNN_{DEC}(y_{t-1},s_{t-1})  
\end{equation}

\subsubsection{Learning}

It is possible to optimize models using different types of training objectives, although maximum log-likelihood (MLE) is the most commonly used method. Given a set of training examples $D =\{(x^s,y^s)\}_{s=1}^{S}$, the MLE is maximised according to Equations \eqref{eqn5} and \eqref{eqn6}.

\begin{equation} \label{eqn5}
\boldsymbol{\hat{\theta}}_{MLE}= \arg \max_{\theta} \{\mathcal{L}(\theta)\}
\end{equation}

\begin{equation} \label{eqn6}
\mathcal{L}(\theta)\ = \sum_{s=1}^{S}logP(y^{s}\vert x^{s});\theta)
\end{equation}

The gradient of $\mathcal{L}$ with respect to $\theta$ is calculated using back-propagation \citep{rumelhart1986learning} as an automatic differentiation algorithm for calculating gradients of the neural network weights, where $\theta$ is the set of model parameters.  

Many NMT approaches implement Stochastic Gradient Descent (SGD) as the optimization algorithm for minimising the loss of the predictive model with regard to the training data. For reasons of computational efficiency, SGD typically computes the loss function and gradients on a minibatch of the training set. The standard SGD optimizer updates parameters of an NMT model according to Equation \eqref{eqn7}, where the learning rate is specified by $\alpha$: 

\begin{equation} \label{eqn7}
\theta \leftarrow \theta - \alpha \bigtriangledown \mathcal{L}(\theta)
\end{equation}
\\
There are several alternatives to using SGD for optimization, among which the ADAM optimizer has proven popular due to a reduction in training times \citep{kingma2014adam}.

\begin{figure}[htp!]
    \centering
    \includegraphics[width=5.5cm]{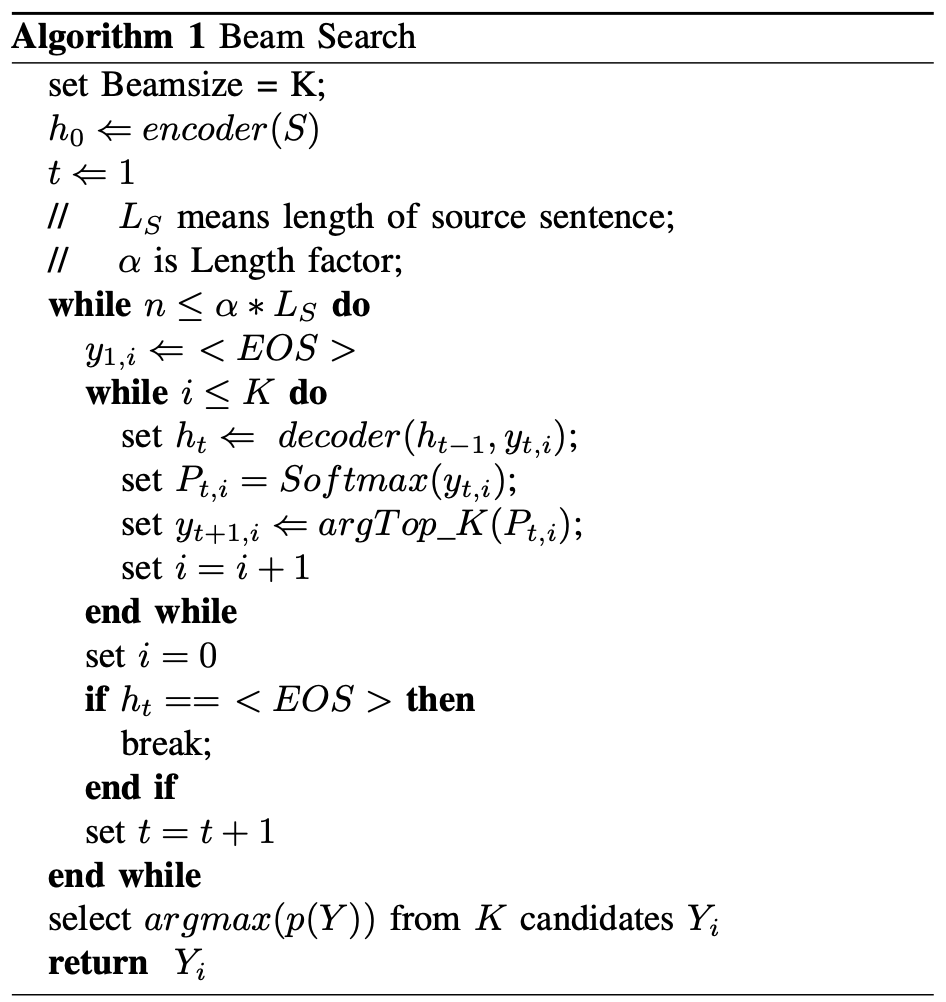}
    \caption{\footnotesize {Beam Search Algorithm} \citep{yang2020survey} }
    \label{fig:beam}
\end{figure}

\subsubsection{Inference}
In the context of NMT, inference should ideally find the target translated sentence $y$ from the source $x$ which maximizes the model prediction $P(y \vert x;\theta)$. However, in practice it is often difficult to find the translation with the highest probability due to the impractically large search space. Accordingly, to find a good but not necessarily the very `best' (i.e. that with the highest probability given the model) translation, NMT usually relies instead on local search algorithms such as greedy search or beam search (cf. Figure \ref{fig:beam}). Translations are carried out by default using beam search, although the option exists to switch to greedy search if needed. This approach is consistent with many other NMT tools since beam search is a classic local search algorithm. Using a pre-defined beam width parameter K, the beam search algorithm keeps only the top-K possible translations as potential candidates.  With each iteration, a new potential translation is formed by combining each candidate word with a new word. New candidate translations compete with each other using log probability values to obtain the new top-K most probable results. This process is continued until the end of the translation process, and the 1-best translation is output.

\subsection{Subword Models}
\label{section:subword}
Translation by its very nature requires an open vocabulary, but restricted (e.g. 30k, 50k, or 70k) vocabularies are typically used for reasons of computational efficiency.  However, the use of subword models aims to address this fixed vocabulary problem associated with NMT. The problem manifests itself in how previously unseen `out-of-vocabulary' (OOV) words are handled. In such cases, a single `UNK' (for `unknown') token is used to `recognize' the OOV word. Encoding rare and unknown words into sequences of subword units significantly reduces the problem and has thus given rise to a number of subword algorithms.

Optimally, this will be performed via morphological processing \citep{passban-etal-2018-tailoring}, but good quality wide-coverage morphological analysers are not always available. Therefore it is common practice to use methods such as Byte Pair Encoding (BPE)~\citep{gage1994new} to break down rare and previously unseen words into subword models in order to significantly improve translation performance~\citep{sennrich2015neural,kudo2018subword}. 

Designed for NMT, SentencePiece~\citep{kudo2018SentencePiece}, is a language-independent subword tokenizer that provides an open-source C++ and a Python implementation for subword units. An attractive feature of the tokenizer is that SentencePiece trains subword models directly from raw sentences.

\subsection{NMT Tools}
\citet{kreutzer2019joey} describe their Joey NMT platform\footnote{\url{https://github.com/joeynmt/joeynmt}} as a minimalist NMT toolkit, based on PyTorch, which is designed especially for newcomers to the field. Joey NMT provides many popular NMT features in a simple code base enabling novice users to easily adapt the system to their particular requirements. The toolkit supports both RNN and Transformer architectures.

Given that adaptNMT is essentially an IPython wrapper layered on top of OpenNMT, it inherits all of OpenNMT’s features and continues to benefit from the work which goes into developing and maintaining its code base. adaptNMT offers a higher level of abstraction over OpenNMT where the focus is much more on usability, especially to newcomers to the field. Accordingly, it provides for easy and rapid deployment by enabling new features  such as greater pre-processing, as well as GUI control over model building. It also contains green features in line with the current research drive towards smaller models with lower carbon footprints (cf. Sections~\ref{sec:envimp} and \ref{disc}). Such features make adaptNMT suitable for both educational and research environments. The key features differentiating adaptNMT from Joey NMT are outlined in Table \ref{tab:joey}.

\begin{table}[]
\begin{tabular}{@{}l@{}}
\toprule
adaptNMT is built upon OpenNMT and subsequently inherits all of its features. \\ \midrule
\begin{tabular}[c]{@{}l@{}}The interface is designed and fully implemented in Google Colab.\end{tabular} \\ \midrule
\begin{tabular}[c]{@{}l@{}}Colab is easier to follow for practitioners since each step can be executed individually.\\ The approach is ideal in education since progression of the  pipeline is demonstrated.\end{tabular} \\ \midrule
Training of models can be viewed and controlled using Colab Android or Apple apps. \\ \midrule
\begin{tabular}[c]{@{}l@{}}adaptNMT can be run in local mode enabling existing infrastructure to be utilised or in \\ hosted mode which allows rapid scaling of the infrastructure.\end{tabular} \\ \midrule
\begin{tabular}[c]{@{}l@{}}Colab Pro+ provides individual researchers, or even small teams, the capacity to build \\ large models on an excellent infrastructure with very little resources.\end{tabular} \\ \midrule
\begin{tabular}[c]{@{}l@{}}GUI controls can split a corpus into train, validation and test datasets. \end{tabular} \\ \midrule
GUI controls are available for hyperparameter customization in NMT training. \\ \midrule
A green report outlines the country-specific kgCO\textsubscript2 generated when training a model.  \\ \midrule
Autonotification notifies the user on completion of training. \\ \midrule
A deploy function enables the immediate deployment of trained models. \\ \midrule
The functionality of serverNMT is not available within Joey NMT. \\ \bottomrule
\end{tabular}
\caption{Key features differentiating adaptNMT from Joey NMT}
\label{tab:joey}
\end{table} 
Other popular frameworks for NMT system-building include FAIRSEQ\footnote{\url{https://github.com/facebookresearch/fairseq}}~\citep{ott2019fairseq}, an open-source sequence modelling toolkit based on PyTorch, that enables researchers to train models for translation, summarization and language modelling. Marian\footnote{\url{https://marian-nmt.github.io}}~\citep{junczys2018marian}, developed using C++, is an NMT framework based on dynamic computation graphs. OpenNMT\footnote{\url{https://opennmt.net}}~\citep{klein2017opennmt} is an open-source NMT framework that has been widely adopted in the research community. The toolkit covers the entire MT workflow from the preparation of data to live inference. 

\subsection{Hyperparameter Optimization}

Hyperparameters are employed in order to customize machine learning models such as translation models. It has been shown that machine learning performance may be improved through hyperparameter optimization (HPO) rather than just using default settings~\citep{sanders2017informing}.

The principal methods of HPO are Grid Search~\citep{montgomery2001design} and Random Search~\citep{bergstra2012random}. Grid search is an exhaustive technique which evaluates all parameter permutations. However, as the number of features grows, the amount of data permutations grows exponentially making optimization expensive in the context of developing translation models which require long build times. Accordingly, an effective, less computationally intensive alternative is to use random search which samples random configurations.

\begin{figure}[ht]
    \centering
    \includegraphics[width=12cm]{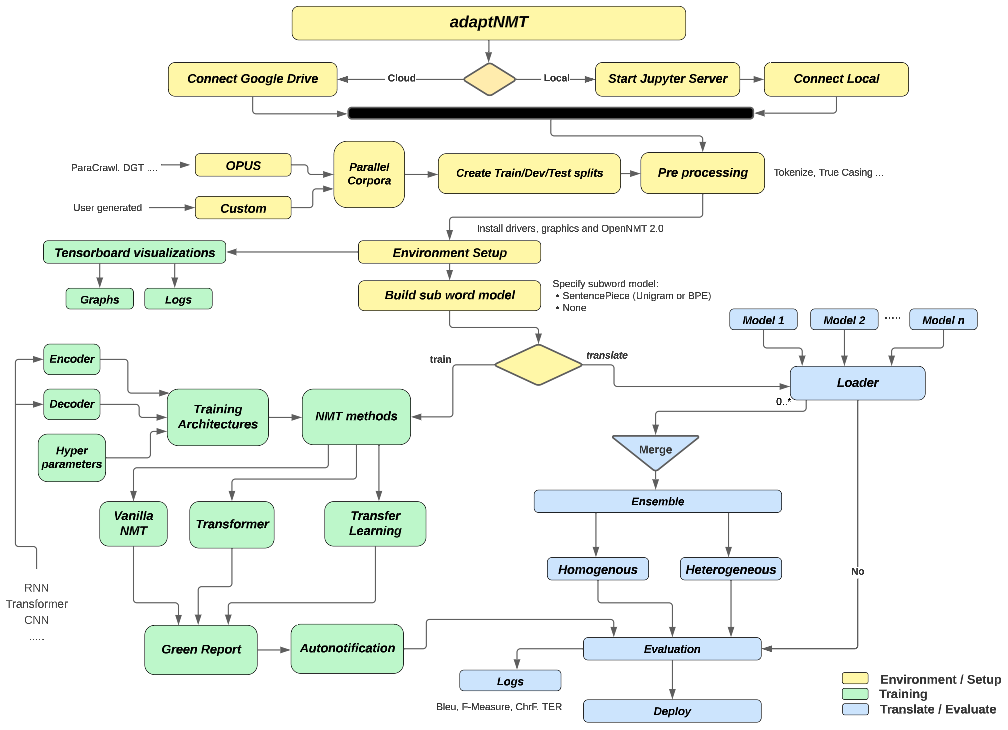}
    \caption{\footnotesize {Proposed architecture for adaptNMT: a language-agnostic NMT development environment. The system is designed to run either in the cloud or using local infrastructure. Models are trained using parallel corpora. Visualization and extensive logging enable real-time monitoring. Models are developed using vanilla RNN-based NMT, Transformer-based approaches or (soon) transfer learning using a fine-tuning approach. Translation and evaluation can be carried out using either single models or ensembles.}}
    \label{fig:approach}
\end{figure}

\section{Architecture of adaptNMT}\label{arch}

Having described the individual components of RNN- and Transformer-based NMT systems, we now present the adaptNMT tool itself, in which these components can be configured by the user. A high-level view of the system architecture of the platform is presented in Figure \ref{fig:approach}. Developed as an IPython notebook, the application uses the Pytorch implementation of \textit{OpenNMT} for training models with SentencePiece used for training subword models. By using a Jupyter notebook, the application may be easily shared with others in the MT community. Furthermore, the difficulties involved in setting up the correct development environment have largely been removed since all required packages are downloaded on-the-fly as the application runs. 

\begin{figure} [htp!]
\centering
\begin{subfigure}{.45\textwidth}
  \includegraphics[width=.9\linewidth]{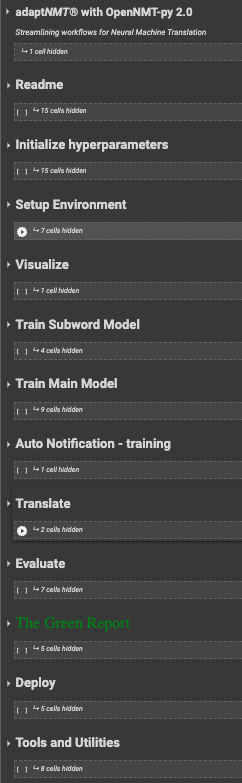}
  \caption{\footnotesize {Overview of adaptNMT. Key areas include initialization, pre-processing, environment setup, visualization, auto and custom NMT, training of subword model, training of main model, evaluation and deployment (cf. Section \ref{aNMT}).} }
  \label{fig:sub1}
\end{subfigure}\hspace{5mm}%
\begin{subfigure}{.45\textwidth}
  \includegraphics[width=.9\linewidth]{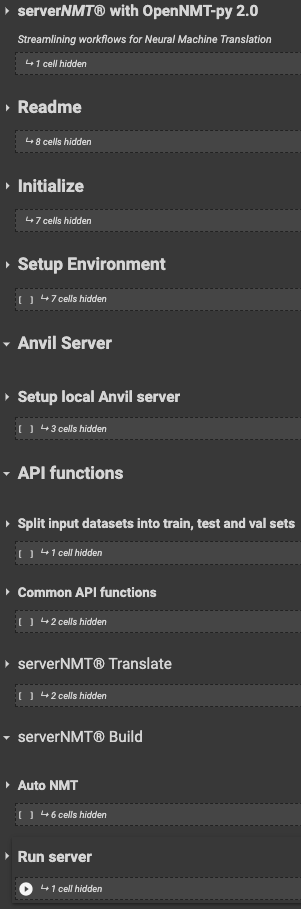}
  \caption{\footnotesize {Overview of serverNMT. Highlighted cells include initialization, environment setup, Anvil server, API functions, translation, model building, adaptNMT and running the server (cf. Section \ref{sNMT}).}}
  \label{fig:sub2}
\end{subfigure}
\caption{adaptNMT and serverNMT}
\label{fig:notebookcells}
\end{figure}

There are options to run the system on local infrastructure or to run it as a Colab instance using Google Cloud. Translation models are developed using parallel text corpora of the source and target languages. A Tensorboard visualization provides a real-time graphical view of model training. The primary use-cases for the system are model building and a translation service, one or both of which can be selected at run-time. As illustrated in the system diagram in Figure \ref{fig:approach}, generating an ensemble output while translating has also been facilitated. Models may also be deployed to a pre-configured location. 

\subsection{adaptNMT}\label{aNMT} 
The application may be run as an IPython Jupyter notebook or as a Google Colab application. Given the ease of integrating large Google drive storage into Colab, the application has been used exclusively as a Google Colab application for our own experiments, some of which are described in Section~\ref{sec:exp}. The key features of the notebook are illustrated in Figure \ref{fig:notebookcells}.

\subsubsection{Initialization and logging}

Initialization enables connection to Google Drive to run experiments, automatic installation of Python, OpenNMT, SentencePiece, Pytorch and other applications. The visualization section enables real-time graphing of  model development.  All log files are stored and can be viewed to inspect training convergence, the model’s training and validation accuracy, changes in learning rates and cross entropy.

\subsubsection{Modes of operation}
There are two modes of operation: local or cloud. In local mode, the application is run so that models are built using the user's local GPU resources. The option to use cloud mode enables users to develop models using Google's GPU clusters. For shorter training times, the unpaid Colab option is adequate. However, for a small monthly subscription, the Google Colab Pro option is worthwhile since users have access to improved GPU and compute resources. Nevertheless, there are also environmental and running costs to consider (cf. Sections~\ref{sec:envimp} and \ref{disc}), although the Google Cloud is run on a platform which uses 100\% renewables \citep{lacoste2019quantifying}. It is also a very cost-effective option for those working in the domain of low-resource languages since developing smaller models require shorter training times. However, users requiring long training times and very high compute resources will need to use their own hardware and run the application in local mode unless they have access to large budgets.

\subsubsection{Customization of models}
The system has been developed to allow users to select variations to the underlying model architecture. A vanilla RNN or Transformer approach may be selected to develop the NMT model. The customization mode enables users to specify the exact parameters required for the chosen approach. One of the features, AutoBuild, enables a user to build an NMT model in three simple steps: (i) upload source and target files, (ii) select  RNN or Transformer, and (iii) click AutoBuild.

\subsubsection{Use of subword segmentation}
The type of optimizer to be used for learning can be specified, and users may also choose to employ different types of subword models when building the system. The subword model functionality allows the user to choose whether or not to use a subword model. Currently, the user specifies the  vocabulary size and chooses either a SentencePiece unigram or a SentencePiece BPE subword model (cf. Section \ref{section:subword}). 

A user may upload a dataset which includes the train, validation and test splits for both source and target languages. In cases where a user has not already created the required splits for model training, single source and target files may be uploaded. The splits needed to create the train, validation and test files are then automatically generated according to the user-specified split ratio. Given that building NMT models typically demands long training times, an automatic notification feature is incorporated that informs the user by email when model training has been completed.

\subsubsection{Translation and evaluation}
In addition to supporting training of models, the application also allows for translation and evaluation of model performance. Translation using pre-built models is also parameterized. Users specify the name of the model as a hyperparameter which is then subsequently used to translate and evaluate the test files. The option for creating an ensemble output is also catered for, and users simply name the models which are to be used in generating the ensemble output.

Once the system has been built, the model to be used for translating the test set may be selected. To evaluate the quality of translation, humans usually provide the best insight, but they may not always be available, do not always agree, and are expensive to recruit for experiments. Accordingly, automatic evaluation metrics are typically used, especially by developers monitoring incremental progress of systems (cf. \citet{Way2018} for more on the pros and cons of human and automatic evaluation).

Several automatic evaluation metrics provided by SacreBleu\footnote{\url{https://github.com/mjpost/sacrebleu}} \citep{post2018call} are used: BLEU \citep{papineni2002bleu}, TER \citep{snover2006study} and ChrF \citep{popovic2015chrf}. Translation quality can also be evaluated using Meteor~\citep{denkowski2014meteor} and F1 score~\citep{melamed-etal-2003-precision}. Note that BLEU, ChrF, Meteor and F1 are precision-based metrics, so higher scores are better, whereas TER is an error-based metric and lower scores indicate better translation quality. Evaluation options available include standard (truecase) and lowercase BLEU scores, a sentence-level BLEU score option, ChrF1 and ChrF3.  

There are three levels of logging for model development, training and experimental results. A references section outlines resources which are relevant to developing, using and understanding adaptNMT. Validation during training is currently conducted using model accuracy and perplexity (PPL). 

\subsection{serverNMT}\label{sNMT}

A  server application, serverNMT, was also developed and implemented as an IPython notebook. It can be configured to run either as a translation server or as a build server. A secure connection, implemented from serverNMT, can be made to websites hosting embedded web apps. At the core of serverNMT, there are two embedded Python web apps, one for translation services and another for developing models, both of which use the anvil.works platform.\footnote{\url{https://anvil.works}} 

As a build server, serverNMT enables a window to the underlying cloud infrastructure in which NMT models can be trained. A web app hosted on another system may connect to this infrastructure made available by serverNMT.

Using an Anvil server embedded within serverNMT, the application continuously waits for communication to web apps and effectively enables a cloud infrastructure for NMT. Written as a REST server, it acts as an API for serving previously built models and facilities the integration of translation models with other systems.

\section{Empirical Evaluation}
\label{sec:exp}

Having described the theoretical background and the tool itself, we now evaluate the effectiveness of the adaptNMT approach by training models for English-Irish (EN-GA) and Irish-English (GA-EN) translation in the health domain using the {\em gaHealth} \citep{lankford2022lrec} corpus.\footnote{\url{https://github.com/seamusl/gaHealth}} All experiments involved concatenating source and target corpora to create a shared vocabulary and a shared SentencePiece subword model. To benchmark the performance of our models, the EN-GA and GA-EN test datasets from the LoResMT2021 Shared Task\footnote{\url{https://github.com/loresmt/loresmt-2021}}~\citep{ojha2021findings} were used. These test datasets enabled the evaluation of the {\em gaHealth} models since the shared task focused on an application of the health domain, namely the translation of Covid-related data. Furthermore, using an official test dataset from a shared task enables the direct comparison of our models' performance with models entered by other teams, as well as future implementations. 

The hyperparameters used for developing the models are outlined in Table \ref{tab:hpo-table}. The details of the train, validation and test sets used by our NMT models are outlined in Tables \ref{tab:en2ga-stats} and  \ref{tab:ga2en-stats}. In all cases, 502 lines were used from the LoResMT2021 validation dataset whereas the test dataset used 502 lines for EN-GA translation and 250 lines for GA-EN translation. Both were independent health-specific Covid test sets which were provided by LoResMT2021. There was one exception; due to a data overlap between the test and train datasets, a reduced test set was used when testing the {\em gaHealth} en2ga* system.  

The results from the IIITT~\citep{puranik2021attentive} and UCF~\citep{chen2021ucf} teams are included in Tables \ref{tab:en2ga} and \ref{tab:ga2en} so  the performance of the {\em gaHealth} models can be easily compared with the findings of the participating LoResMT2021 systems. IIITT fine-tuned an Opus MT model\footnote{\url{https://github.com/Helsinki-NLP/Opus-MT}}~\citep{tiedemann-thottingal-2020-opus} on the training dataset.  UCF used transfer learning \citep{zoph-etal-2016-transfer}, unigram and subword segmentation methods for EN–GA and GA–EN translation.

\begin{center}
\begin{table}
\center
\begin{tabular}{ll}
\hline
\textbf{Hyperparameter} & \textbf{Values}                \\ \hline
Learning rate            & 0.1, 0.01, 0.001, \textbf{2}            \\ \hline
Batch size               & 1024, \textbf{2048},  4096, 8192       \\ \hline
Attention heads          & \textbf{2}, 4, \textbf{8}                     \\ \hline
Number of layers         & 5, \textbf{6}                           \\ \hline
Feed-forward dimension   & \textbf{2048}                           \\ \hline
Embedding dimension      & 128, \textbf{256}, 512                  \\ \hline
Label smoothing          & \textbf{0.1}, 0.3                       \\ \hline
Dropout                  & 0.1, \textbf{0.3}                       \\ \hline
Attention dropout        & \textbf{0.1}                            \\ \hline
Average Decay            & 0, \textbf{0.0001}                      \\ \hline
\end{tabular}
\caption{Hyperparameter Optimization for Transformer models. Optimal parameters are highlighted in bold \citep{lankford2021transformer}.}
\label{tab:hpo-table}
\end{table}
\end{center}

\begin{table} [ht]
\centering
\begin{tabular}{lcccccc}
\hline
\textbf{Team} &
 \textbf{System} &
  \textbf{Train}  &
  \textbf{Validation}  &
  \textbf{Test}  \\ \hline
adapt & covid\_extended & 13k & 502 & 500 \\
adapt & combined\_domains & 65k & 502 & 500 \\
IIITT  & en2ga-b & 8k & 502 & 500 \\
UCF     & en2ga-a & 8k & 502 & 500   \\ 
{\em gaHealth} & en2ga & 24k & 502 & 500 \\
{\em gaHealth} & en2ga* & 24k & 502 & 338 \\
\hline
\end{tabular}
\caption{EN-GA train, validation and test dataset distributions. The baseline {\em gaHealth} system was augmented with an 8k Covid dataset provided by LoResMT2021.} 
\label{tab:en2ga-stats}
\end{table}

\begin{table} [ht]
\centering
\begin{tabular}{lcccccc}
\hline
\textbf{Team} &
 \textbf{System} &
  \textbf{Train}  &
  \textbf{Validation}  &
  \textbf{Test}  \\ \hline
IIITT  & ga2en-b & 8k & 502 & 250 \\
UCF     & ga2en-b & 8k & 502 & 250   \\ 
{\em gaHealth} & ga2en & 24k & 502 & 250 \\
\hline
\end{tabular} 
\caption{GA-EN train, validation and test dataset distributions. The baseline {\em gaHealth} system was augmented with an 8k Covid dataset provided by LoResMT2021. All overlaps were removed from the {\em gaHealth} corpus prior to training the {\em gaHealth} ga2en model.}
\label{tab:ga2en-stats}
\end{table}

\subsection{Infrastructure}
Rapid prototype development was enabled through a Google Colab Pro subscription using NVIDIA Tesla P100 PCIe 16GB graphic cards and up to 27GB of memory when available~\citep{bisong2019google}. All {\em gaHealth} MT models were trained using {\em adaptNMT}. 

\subsection{Metrics}

Automated metrics were used to determine the translation quality. In order to compare against our previous work, the performance of models is measured using three evaluation metrics, namely BLEU, TER and ChrF. These metrics indicate the accuracy of the translations derived from our NMT systems.

Case-insensitive BLEU scores at the corpus level are reported. Model training was stopped after 40k training steps or once an early stopping criterion of no improvement in validation accuracy for four consecutive iterations was recorded.

PPL is often used to evaluate language models within NLP. It measures the effectiveness of a probability model in predicting a sample. As a metric for translation performance, it is important to keep low scores so that the number of alternative translations is reduced.

\subsection{Results: Automatic Evaluation}

\begin{figure}[htbp]
    \centering
    {\includegraphics[width=3.95cm]{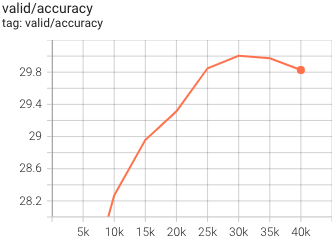}}
    {\includegraphics[width=3.95cm]{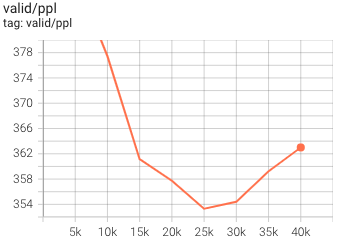}}
    \caption{ \footnotesize {adapt covid\_extended system: training \textit{EN-GA} model with 13k lines consisting of the ADAPT 5k corpus and an 8k LoResMT2021 Covid corpus. The graph on the left illustrates OpenNMT accuracy and the graph on the right demonstrates perplexity.}}
    \label{fig:en-ga-covid}
\end{figure}

The experimental results from LoResMT 2021 are summarized in Tables \ref{tab:en2ga} and \ref{tab:ga2en}. In the LoResMT2021 Shared Task, the highest-performing EN-GA system was submitted by the ADAPT team \citep{lankford2021machine}. The system uses an extended Covid dataset, which is a combination of the 2021 MT Summit Covid baseline and a custom ADAPT Covid dataset. The model, developed within adaptNMT, uses a Transformer architecture with 2 heads. It performs well across all key translation metrics (BLEU: 36.0, TER: 0.531 and ChrF3: 0.6).The training of this EN-GA model is illustrated in Figure \ref{fig:en-ga-covid}.The model achieved a maximum validation accuracy of 30.0\% and perplexity of 354 after 30k steps. 

\begin{table}[ht!]
\centering
\begin{tabular}{lcccccc}
\hline
\textbf{Team} &
 \textbf{System} &
  \textbf{BLEU} $\uparrow$ &
  \textbf{TER} $\downarrow$ &
  \textbf{ChrF3} $\uparrow$ \\ \hline
UCF     & en2ga-b & 13.5 & 0.756 & 0.37   \\
IIITT  & en2ga-b & 25.8 & 0.629 & 0.53 \\
adapt & combined & 32.8 & 0.590 & 0.57 \\
{\em gaHealth} & en2ga & 33.3 & 0.604 & 0.56 \\
adapt & covid\_extended & 36.0 & 0.531 & 0.60 \\
{\em gaHealth} & en2ga* & \textbf{37.6} & 0.577 & 0.57 \\  
\hline
\end{tabular}
\caption{EN-GA {\em gaHealth} system compared with LoResMT 2021 EN-GA systems.}
\label{tab:en2ga}
\end{table}

\begin{figure}[ht!]
    \centering
    {\includegraphics[width=3.95cm]{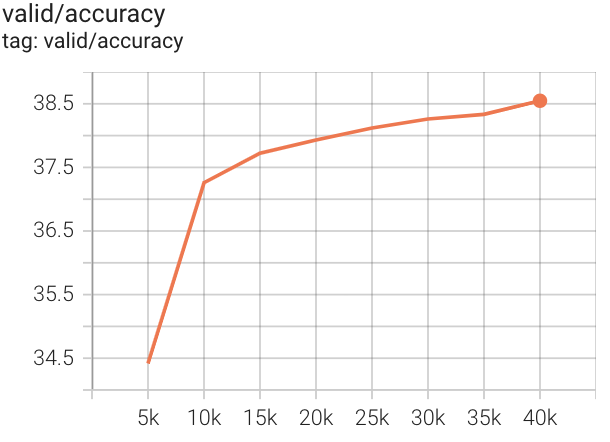}}
    {\includegraphics[width=3.95cm]{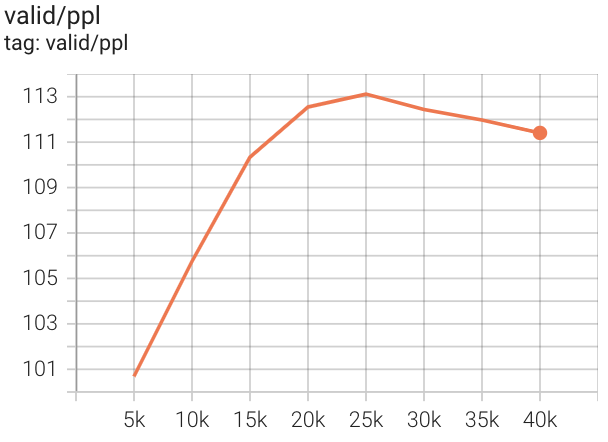}}
    \caption{\footnotesize {{\em gaHealth} en2ga* system: training \textit{EN-GA} model with combined 16k gaHealth corpus and 8k LoResMT2021 Covid corpus. The graph on the left illustrates OpenNMT accuracy and the graph on the right demonstrates perplexity.}}
    \label{fig:en-ga-gaHealth}
\end{figure}

\begin{figure}[h]
    \centering
    {\includegraphics[width=3.95cm]{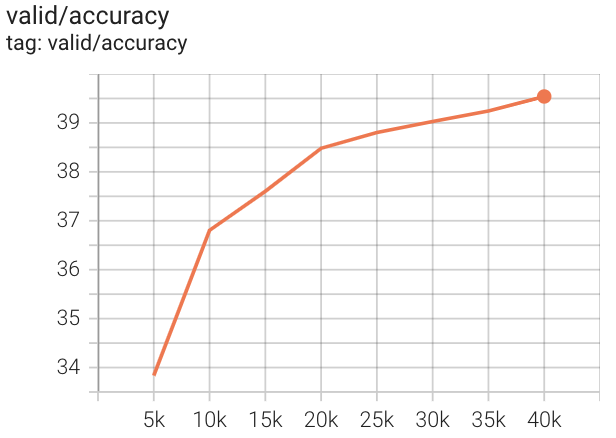}}
    {\includegraphics[width=3.95cm]{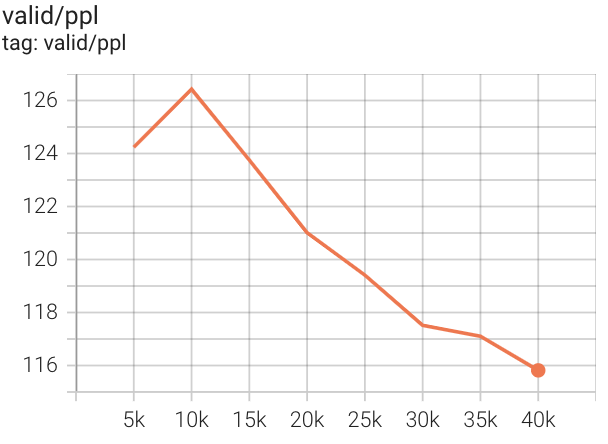}}
    \caption{\footnotesize { {\em gaHealth} ga2en system: training \textit{GA-EN} model with combined 16k gaHealth corpus and 8k LoResMT2021 Covid corpus. The graph on the left illustrates OpenNMT accuracy and the graph on the right demonstrates perplexity. }}
    \label{fig:ga-en-gaHealth}
\end{figure}

The results from the LoResMT2021 Shared Task were further improved by developing models using a bespoke health dataset, {\em gaHealth}. Table \ref{tab:en2ga} shows an improvement of 1.6 BLEU points, a relative improvement of almost 4.5\%, although TER and ChrF3 scores are a little worse. Validation accuracy and PPL in training the {\em gaHealth} models with adaptNMT are illustrated in Figures \ref{fig:en-ga-gaHealth} and \ref{fig:ga-en-gaHealth}. Figure \ref{fig:en-ga-covid} illustrates model training using the covid\_extended dataset, also developed using adaptNMT. In training the {\em gaHealth} en2ga* system, as highlighted in Figure \ref{fig:en-ga-gaHealth}, the EN-GA model was trained with the combined 16k {\em gaHealth} and 8k LoResMT2021 corpora. The model's validation accuracy of 38.5\% and perplexity of 113 achieved a BLEU score of 37.6 when evaluated with the test data.

The training of the GA-EN {\em gaHealth} ga2en system with the combined 16k gaHealth corpus and 8k LoResMT2021 Covid corpus is shown in Figure \ref{fig:ga-en-gaHealth}. This model achieves a validation accuracy of 39.5\% and perplexity of 116 which results in a BLEU score of 57.6. This is significantly better (by 20 BLEU points) than for the reverse direction, as it is well-known that translating into a morphologically-rich language like Irish is always more difficult compared to when the same language acts as the source. This is confirmed by comparing the results for the UCF (13.5 vs. 21.3 BLEU) and IIITT (25.8 vs. 34.6) systems in Tables~\ref{tab:en2ga} and \ref{tab:ga2en}.

Rapid convergence was observed while training the {\em gaHealth} models such that little accuracy improvement occurs after 30k steps, 10K fewer than for the reverse direction. Only marginal gains were achieved after this point and it actually declined in the case of the system trained using the covid\_extended dataset, as the left-hand graph in Figure~\ref{fig:en-ga-covid} shows. 

\par
Of the models developed by the ADAPT team, the worst-performing model uses a larger 65k dataset. This is not surprising given that the dataset is from a generic domain of which only 20\% is health related. The performance of this higher-resourced 65k line model lags behind the augmented {\em gaHealth} model which was developed using just 24k lines. 

\begin{table}[ht!]
\centering
\begin{tabular}{lcccccc}
\hline
\textbf{Team} &
 \textbf{System} &
  \textbf{BLEU} $\uparrow$ &
  \textbf{TER} $\downarrow$ &
  \textbf{ChrF3} $\uparrow$ \\ \hline
UCF & ga2en-b & 21.3 & 0.711 & 0.45\\
IIITT  & ga2en-b & 34.6 & 0.586 & 0.61\\
{\em gaHealth} & ga2en & \textbf{57.6} & 0.385 & 0.71\\
\hline
\end{tabular} 
\caption{GA-EN {\em gaHealth} systems compared with LoResMT 2021 GA-EN systems.}
\label{tab:ga2en}
\end{table}

\par
For translation in the GA-EN direction, the best-performing model for the LoResMT2021 Shared Task was developed by IIITT with a BLEU of 34.6, a TER of 0.586 and ChrF3 of 0.6. Accordingly, this serves as the baseline score by which our GA-EN model, developed using the {\em gaHealth} corpus, can be benchmarked. The performance of the {\em gaHealth} model offers an improvement across all metrics with a BLEU score of 57.6, a TER of 0.385 and a ChrF3 result of 0.71. In particular, the 66\% relative improvement in BLEU score against the IIITT system is very significant.

\subsection{Environmental Impact}
\label{sec:envimp}
We were motivated by the findings of \citet{strubell-etal-2019-energy} and \citet{bender2021dangers} to track the energy consumption required to train our models. Prototype model development used Colab Pro, which as part of Google Cloud is carbon neutral \citep{lacoste2019quantifying}. However, longer running Transformer experiments were conducted on local servers using 324 gCO\textsubscript2 per kWh\footnote{\url{https://www.seai.ie/publications/Energy-in-Ireland-2020.pdf}}\citep{sei2020}. The net result was just under 10 kgCO\textsubscript2 created for a full run of model development. Models developed during this study will be reused for ensemble experiments in the future so that work will have a life beyond this paper.

\subsection{Stochastic Nuances}
\label{sec:stoc}

To evaluate the translation performance of an IPython-based application such as adaptNMT, a comparison with a Python script version of the same application, myNMT.py, was conducted.  We built translation models in the EN-GA and the GA-EN directions using this script. The models developed with adaptNMT were trained on Google Colab using a 12GB Tesla K80 GPU, whereas the myNMT models were trained on a local machine using a 12GB Gigabyte 3060 graphics card. The results from evaluating these models are presented in Tables~\ref{tab:engastoc} and \ref{tab:gaenstoc}. 

Despite setting the same random seed, it is clear from Tables~\ref{tab:engastoc} and \ref{tab:gaenstoc} that the translation performance of the adaptNMT models is better by 1.2 BLEU points (3.3\% relative improvement) in the EN-GA direction and 1.0 BLEU point (1.8\% relative improvement) in the GA-EN direction. 

Given the stochastic nature of machine learning, training models on different systems can give yield different results even with the same train, validation and test data. The performance differences can be attributed to the stochastic nature of the learning algorithm and evaluation procedure. Furthermore the platforms had different underlying system architectures which is another source of stochastic error.

\begin{table}[ht!]
\centering
\begin{tabular}{lcccccc}
\hline
 \textbf{System} &
  \textbf{BLEU} $\uparrow$ &
  \textbf{TER} $\downarrow$ &
  \textbf{ChrF3} $\uparrow$ \\ \hline
adaptNMT & 37.6 & 0.577 & 0.570 \\
myNMT  & 36.4 & 0.622 & 0.56 \\
\hline
\end{tabular} 
\caption{Stochastic differences between EN-GA systems}
\label{tab:engastoc}
\end{table}

\begin{table}[ht!]
\centering
\begin{tabular}{lcccccc}
\hline
 \textbf{System} &
  \textbf{BLEU} $\uparrow$ &
  \textbf{TER} $\downarrow$ &
  \textbf{ChrF3} $\uparrow$ \\ \hline
adaptNMT & 57.6 & 0.385 & 0.71 \\
myNMT  & 56.6 & 0.399 & 0.703 \\
\hline
\end{tabular} 
\caption{Stochastic differences between EN-GA systems}
\label{tab:gaenstoc}
\end{table}

\section{Discussion}\label{disc}
The mathematical first principles governing NMT development were presented to demonstrate the mechanics of what happens during model training. Several parameters in Equations (\ref{eqn1})-(\ref{eqn7}) are configurable  within the adaptNMT application.
 
The environmental impact of technology, and the measurement of its effects, has gained a lot of prominence in recent years \citep{henderson2020towards}. Indeed, this may be viewed as a natural response to truly massive NLP models which have been developed by large multinational corporations.  In particular, HPO of NMT models can be particularly demanding if hyperparameter fine-tuning is conducted across a broad search space. As part of their work on NMT architectures, the Google Brain team required more than 250,000 GPU hours for NMT HPO \citep{britz2017massive}. Training of these models was conducted using Tesla K40m and Tesla K80 GPUs with maximum power consumption between 235W and 300W, giving rise to potentially in excess of 60 MWh of energy usage. Even though the Google Cloud is carbon neutral, one must consider the opportunity cost of this energy usage. 
 
A plethora of tools to evaluate the carbon footprint of NLP \citep{bannour2021evaluating} has subsequently been developed and the concept of sustainable NLP has become an important research track in its own right at many high profile conferences such as the EACL 2021 \textit{Green and Sustainable NLP} track.\footnote{\url{https://2021.eacl.org/news/green-and-sustainable-nlp}} 
In light of such developments, a `green report' was incorporated into adaptNMT whereby the kgCO\textsubscript2 generated during model development is  logged. This is very much in line with the industry trend of quantifying the impact of NLP on the environment; indeed, \citet{info13020088} have demonstrated that high-performing MT systems can be built with much lower footprints, which not only reduce emissions, but also in the post-deployment phase deliver savings of almost 50\% in energy costs for a real translation company.

To evaluate system performance in translating health data in the EN-GA direction, we used the adaptNMT application to develop an MT model for the LoResMT2021 Shared Task. The application was subsequently used to develop an MT model for translating in the GA-EN direction. In both cases, high-performing models achieving SOTA scores were achieved by using adaptNMT to develop  Transformer models capable of generating high-quality output.    

The danger of relying on increasingly large language models has been well-documented in the literature. Such discussion focuses not just on the environmental impact but also highlights the impact of in-built bias and the inherent risks that large models pose for low-resource languages \citep{bender2021dangers}. Using an easily-understood framework such as  adaptNMT, the benefits of developing high-performing NMT models with smaller in-domain datasets should not be overlooked. 

\section{Conclusion and Future Work}\label{concl}

We introduced adaptNMT, an application for NMT which manages the complete workflow of model development, evaluation and deployment. The performance of the application was demonstrated in the context of generating an EN-GA translation model which ranked 1st in the LoResMT2021 shared task, and validated against a standalone reimplementation of both EN-GA and GA-EN systems outside the tool, where no drop-off in performance was seen. 

With regard to future work, development will focus more on tracking environmental costs and integrating new transfer learning methods. Modern zero-shot and few-shot approaches, adopted by GPT3 \citep{brown2020language} and Facebook LASER \citep{artetxe2019massively} frameworks, will be integrated. Whereas the existing adaptNMT application focuses on customizing NMT models, a separate application adaptLLM will be developed to fine-tune large language models, in particular those that focus on low-resource language pairs such as NLLB \citep{costa2022no}.

The green report embedded within the application is our first implementation of a sustainable NLP feature within adaptNMT. It is planned to develop this feature further to include an improved UI and user recommendations about how to develop greener models.  As an open-source project, we hope the community will add to its development by contributing new ideas and improvements. 

\section*{Declarations}

This research is supported by Science Foundation Ireland through ADAPT Centre (Grant 13/RC/2106) (www.adaptcentre.ie) at Dublin City University. This research was also funded by the Munster Technological University and the National Relay Station (NRS) of Ireland.

\bibliography{sn-bibliography}

\end{document}